# Plant Disease Detection Using Image Processing and Machine Learning

Pranesh Kulkarni[1], Atharva Karwande[1], Tejas Kolhe[1], Soham Kamble[1], Akshay Joshi[1], Medha Wyawahare[1]

[1] Department of Electronics and Telecommunication, Vishwakarma Institute of Technology, Pune, India.

**Abstract.** One of the important and tedious task in agricultural practices is detection of disease on crops. It requires huge time as well as skilled labor. This paper proposes a smart and efficient technique for detection of crop disease which uses computer vision and machine learning techniques. The proposed system is able to detect 20 different diseases of 5 common plants with 93% accuracy.

## 1 Introduction

In India about 70% of the populace relies on agriculture. Identification of the plant diseases is important in order to prevent the losses within the yield.  It's terribly troublesome to observe the plant diseases manually. It needs tremendous quantity of labor, expertize within the plant diseases, and conjointly need the excessive time interval. Hence, image processing and machine learning models can be employed for the detection of plant diseases. In this project, we have described the technique for the detection of plant diseases with the help of their leaves pictures. Image processing is a branch of signal processing which can extract the image properties or useful information from the image. Machine learning is a sub part of artificial intelligence which works automatically or give instructions to do a particular task. The main aim of machine learning is to understand the training data and fit that training data into models that should be useful to the people. So it can assist in good decisions making and predicting the correct output using the large amount of training data. The color of leaves, amount of damage to leaves, area of the leaf, texture parameters are used for classification. In this project we have analyzed different image parameters or features to identifying different plant leaves diseases to achieve the best accuracy. Previously plant disease detection is done by visual inspection of the leaves or some chemical processes by experts. For doing so, a large team of experts as well as continuous observation of plant is needed, which costs high when we do with large farms. In such conditions, the recommended system proves to be helpful in monitoring large fields of crops. Automatic detection of the diseases by simply seeing the symptoms on the plant leaves makes it easier as well as cheaper. The proposed solution for plant disease detection is computationally less expensive and requires less time for prediction than other deep learning based approaches since it uses statistical machine learning and image processing algorithm.

## 2   Literature Review

In 2015, S. Khirade et Al. tackled the problem of plant disease detection using digital image processing techniques and back propagation neural network (BPNN) [1]. Authors have elaborated different techniques for the detection of plant disease using the images of leaves. They have implemented Otsu's thresholding followed by boundary detection and spot detection algorithm to segment the infected part in leaf. After that they have extracted the features such as color, texture, morphology, edges etc. for classification of plant disease. BPNN is used for classification i.e. to detect the plant disease.

Shiroop Madiwalar and Medha Wyawahare analyzed different image processing approaches for plant disease detection in their research [2]. Authors analyzed the color and texture features for the detection of plant disease. They have experimented their algorithms on the dataset of 110 RGB images. The features extracted for classification were mean and standard deviation of RGB and YCbCr channels, grey level co-occurrence matrix (GLCM) features, the mean and standard deviation of the image convolved with Gabor filter. Support vector machine classifier was used for classification. Authors concluded that GCLM features are effective to detect normal leaves. Whereas color features and Gabor filter features are considered as best for detecting anthracnose affected leaves and leaf spot respectively. They have achieved highest accuracy of 83.34% using all the extracted features.

Peyman Moghadam et Al. demonstrated the application of hyperspectral imaging in plant disease detection task [3]. visible and near-infrared (VNIR) and short-wave infrared (SWIR) spectrums were used in this research. Authors have used k-means clustering algorithm in spectral domain for the segmentation of leaf. They have proposed a novel grid removal algorithm to remove the grid from hyperspectral images. Authors have achieved the accuracy of 83% with vegetation indices in VNIR spectral range and 93% accuracy with full spectrum. Though the proposed method achieved higher accuracy, it requires the hyperspectral camera with 324 spectral bands so the solution becomes too costly.

Sharath D. M. et Al. developed the Bacterial Blight detection system for Pomegranate plant by using features such as color, mean, homogeneity, SD, variance, correlation, entropy, edges etc. Authors have implemented grab cut segmentation for segmenting the region of interest in the image [4]. Canny edge detector was used to extract the edges from the images. Authors have successfully developed a system which can predict the infection level in the fruit.

Garima Shrestha et Al. deployed the convolutional neural network to detect the plant disease [5]. Authors have successfully classified 12 plant diseases with 88.80% accuracy. The dataset of 3000 high resolution RGB images were used for experimentation. The network has 3 blocks of convolution and pooling layers. This makes the network computationally expensive. Also the F1 score of the model is 0.12 which is very low because of higher number of false negative predictions.

## 3 Methodology

### 3.1 Dataset

For this project we have used public dataset for plant leaf disease detection called PlantVillage curated by Sharada P. Mohanty et Al. [6]. The dataset consists of 87000 RGB images of healthy and unhealthy plant leaves having 38 classes out of which We have selected only 25 classes for experimentation of our algorithm These classes are shown in Table 1.

**Table 1.** Dataset Specifications.

| Plant | Disease Name | No. of Images |
|---|---|---|
| Apple | Healthy | 2008 |
| | Diseased Scab | 2016 |
| | Diseased: Black rot | 1987 |
| | Diseased: Cedar apple rust | 1760 |
| Corn | Healthy | 1859 |
| | Diseased: Cercospora leaf spot | 1642 |
| | Diseased: Common rust | 1907 |
| | Diseased: Northern Leaf Blight | 1908 |
| Grapes | Healthy | 1692 |
| | Diseased: Black rot | 1888 |
| | Diseased: Esca (Black Measles) | 1920 |
| | Diseased: Leaf blight (Isariopsis) | 1722 |
| Potato | Healthy | 1824 |
| | Diseased: Early blight | 1939 |
| | Diseased: Late blight | 1939 |
| Tomato | Healthy | 1926 |
| | Diseased: Bacterial spot | 1702 |
| | Diseased: Early blight | 1920 |
| | Diseased: Late blight | 1851 |
| | Diseased: Leaf Mold | 1882 |
| | Diseased: Septoria leaf spot | 1745 |
| | Diseased: Two-spotted spider mite | 1741 |
| | Diseased: Target Spot | 1827 |
| | Diseased: Yellow Leaf Curl Virus | 1961 |
| | Diseased: Tomato mosaic virus | 1790 |

Some samples from the datset are shown in Fig. 1.

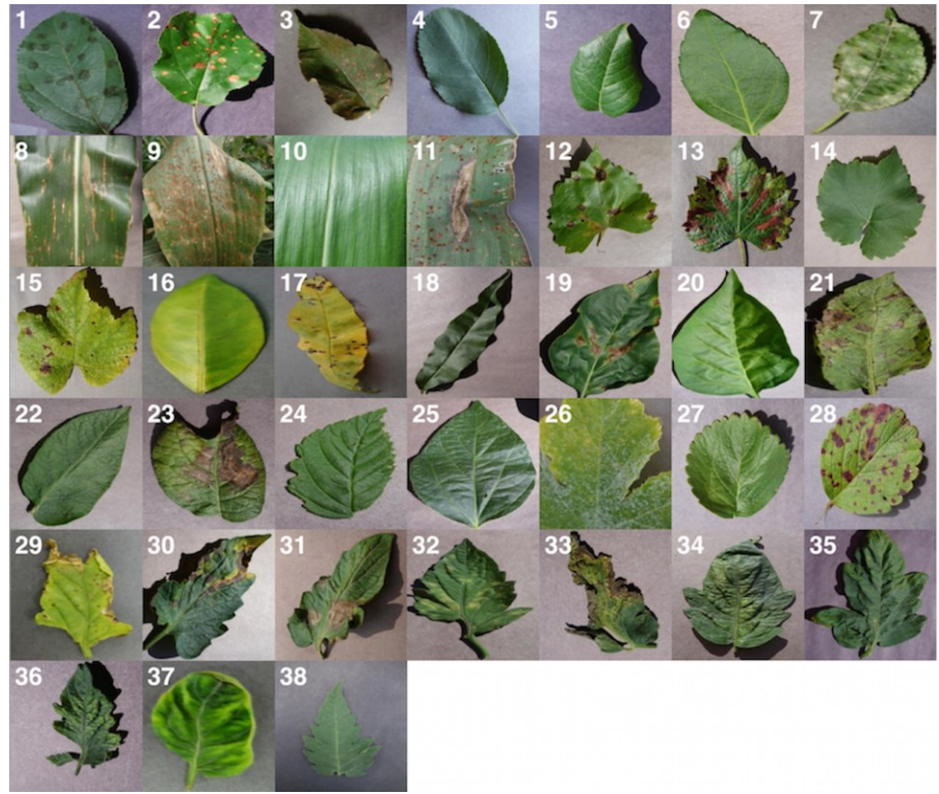

**Fig. 1.** Sample images in the dataset [6].

### 3.2 Data preprocessing and feature extraction

Data preprocessing is important task in any computer vision based system. Fig. 2 illustrates the preprocessing steps for each image. To get precise results, some background noise should be removed before extraction of features. So first the RGB image is converted to greyscale and then Gaussian filter is used for smoothening of the image. Then to binaries the image, Otsu's thresholding algorithm is implemented. Then morphological transform is applied on binarised image to close the small holes in the foreground part. Now after foreground detection, the bitwise AND operation on binarised image and original color image is performed to get RGB image of segmented leaf. Now after image segmentation shape, texture and color features are extracted from the image. By using contours, area of the leaf and perimeter of the leaf is calculated. Contours are the line that joins all the points along the edges of objects having same color or intensity. Mean and standard deviation of each channel in RGB image is also estimated. To obtain amount of green color in the image, image is first converted to HSV color space and we have calculated the ratio of number of pixels having pixel intensity of hue (H) channel in between 30 and 70 and total number of pixels in one channel. Non green part of image is calculated by subtracting green color part from 1.

After extracting color features from the image, we have extracted texture features from grey level co-occurrence matrix (GLCM) of the image [7].

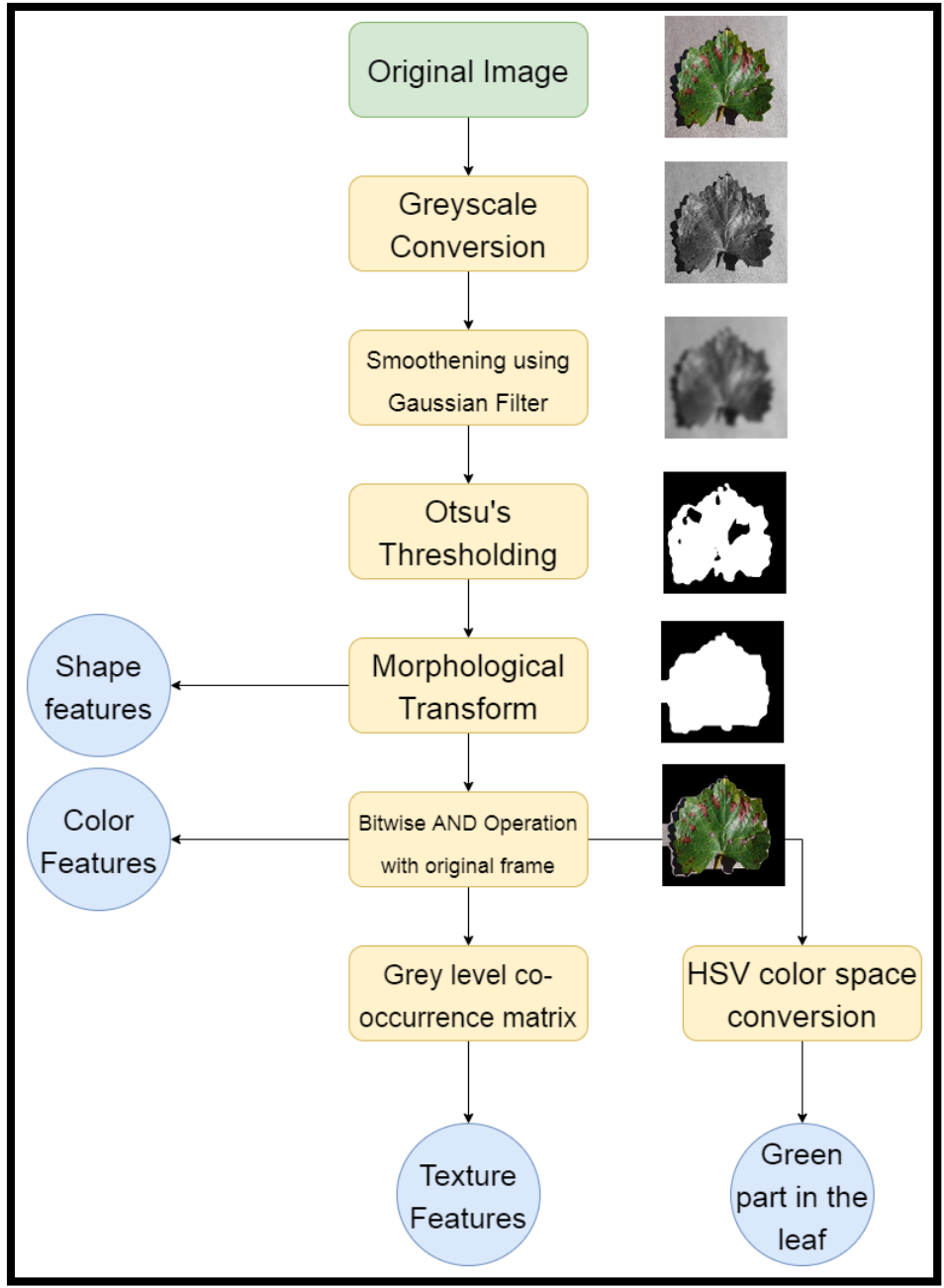

**Fig. 2.** Steps for data preprocessing and feature extraction.

GLCM is the spacial relationship of pixels in the image. Extracting texture features from GCLM is one of the tradition method in computer vision. We have extracted following features from GCLM:

- Contrast
- Dissimilarity
- Homogeneity
- Energy
- Correlation

After extracting all the features from all the images in the dataset, feature selection task is performed.

### 3.3 Feature selection

Feature selection is an important step in all machine learning problems. In this project we are selecting the features on the basis of correlation of variables with target variable. Fig. 3 shows the correlation of each variable with each other for apple dataset. The correlation of feature green part of leaf (F1) and green part of leaf (F2) is very high (1) which means both variables are dependent on each other. So we have dropped one of them (F2). Now for apple disease prediction, less correlated features such as green channel mean, red channel standard deviation, blue channel standard deviation, dissimilarity (f5) and correlation (f8) will not contribute too much in model development. So we have dropped these variables also. After feature selection, the data is now parsed to machine learning classifiers to find the patterns in the data.

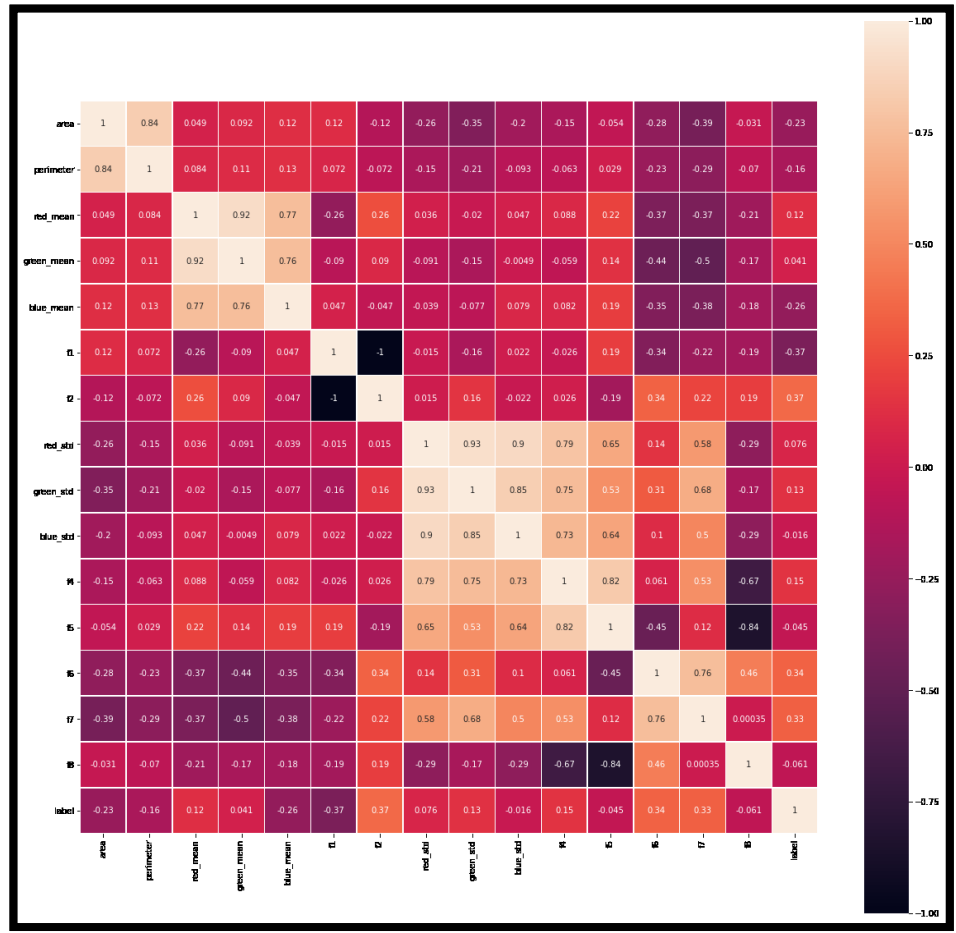

**Fig. 3.** Correlation plot for Apple dataset.

### 3.4 Classification Algorithm

Random forest classifier has been used for classification or detection task. It is the part of ensemble learning, where the output is predicted from multiple base estimators [8]. Generally, to achieve higher accuracies, decision trees are used. But they are prone to overfitting problems. So to overcome this issue, random forest classifier is used which is a combination of multiple decision trees. Each tree is trained by using different subsets of the whole dataset, this can reduce the overfitting and improves the accuracy of the classifier. We have splitted the dataset into train set (80%) for fitting the model and test set (20%) for validation. K-fold cross validation technique is implemented to find the accuracy score. This method can find the accuracy on whole dataset without any bias. After fitting the data, f1 score, precision, recall, accuracy has been calculated from test data to analyze the performance of the model. ROC curve and confusion matrix was plotted to analyze false positives and false negatives.

## 4 Results and discussion

Table 2 shows the performance matrices for each model developed for each of the plant. We can observe that the accuracy scores are nearly equal to f1 scores. This is because of balanced number of false negative and false positive predictions. This is considered as best case for any machine learning algorithm. The average accuracy was 93%.

**Table 2.** Performance matric for all models.

| Plant | Accuracy | F1 Score |
| --- | --- | --- |
| Apple | 0.91 | 0.91 |
| Corn | 0.94 | 0.94 |
| Grapes | 0.95 | 0.95 |
| Potato | 0.98 | 0.98 |
| Tomato | 0.87 | 0.87 |

Fig. 4 shows the confusion matrices for each of the model. With the help of confusion matrices, number of false negatives, false positives, true predictions can be analyzed. Fig. 5 shows the receiver operating characteristic (ROC) curve for each of the model. An ROC curve is a graph showing the performance of a classification model at all classification thresholds. It depends upon two parameters, true positive rate and false positive rate.

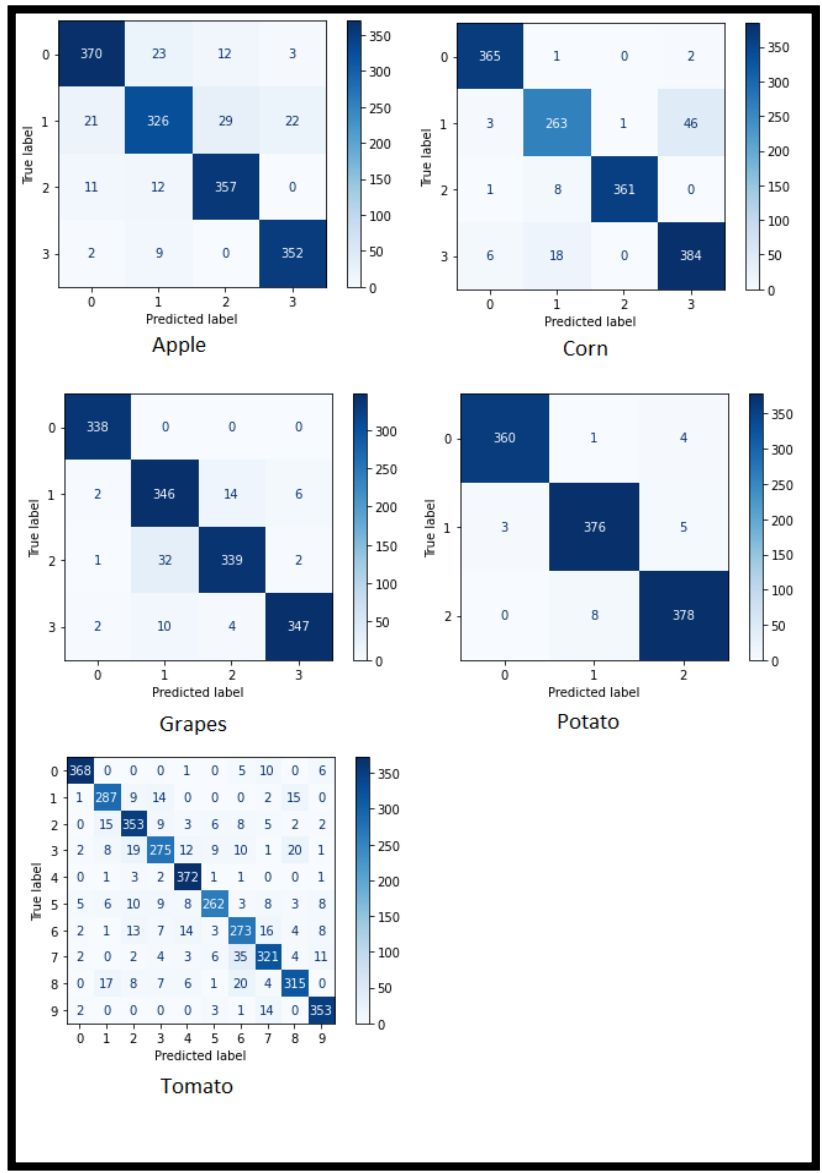

**Fig. 4.** Confusion matrices for all the models.

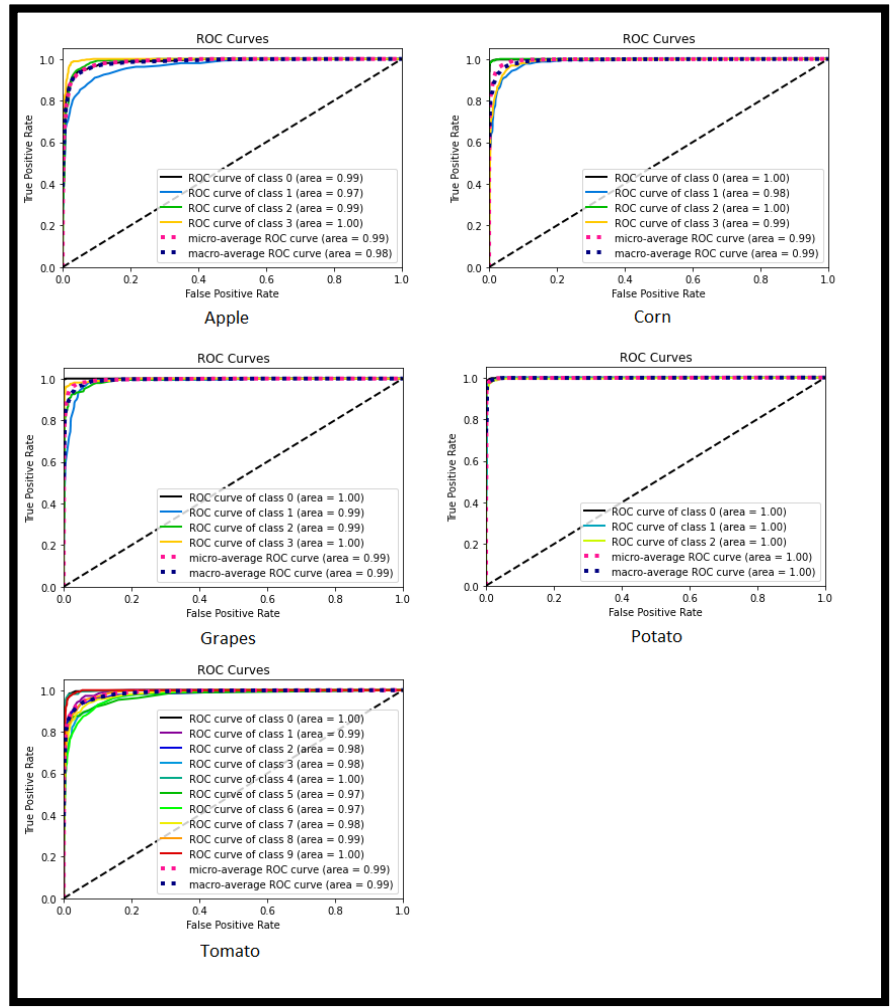

**Fig. 5.** ROC curves for all the models.

We have developed a flask based web application for detecting the plant disease and deployed it on heroku (free cloud hosting server). Fig 6 shows the homepage of deployed web application and Fig 7 shows the input images and their corresponding predictions made by our system. It shows that the system successfully detected the disease of leaf.

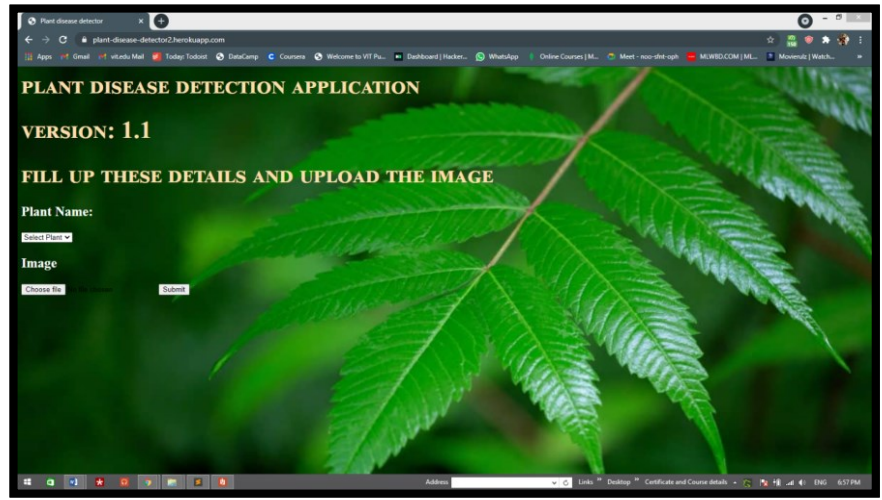

**Fig. 6.** Homepage of deployed API.

However, we can deploy an intelligent robot vehicle with high end processor attached to it for real time plant disease detection. This system can detect the diseased plants in the agricultural site. Even we can automate the process of spreading the fertilizers by using such robots. Our proposed algorithm is computationally inexpensive, so it can detect the plant disease in efficient manner. Also sometimes it happens that the farmer also could not identify the disease of the plant. So they need an expert advice. So we can deploy a website which can detect the plant disease based on images captured and uploaded by farmer and can give suggestions or can suggest some fertilizers based on detected disease.

| Input Image | Predictions |
|---|---|
| 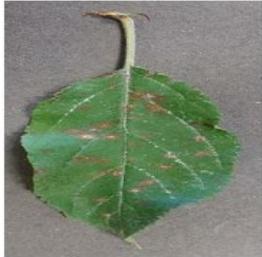 | 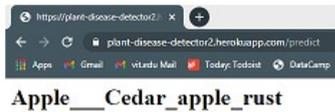 Apple___Cedar_apple_rust |
| 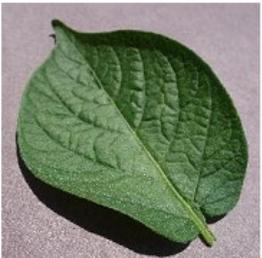 | 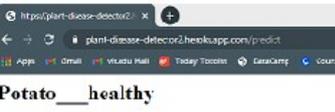 Potato___healthy |
| 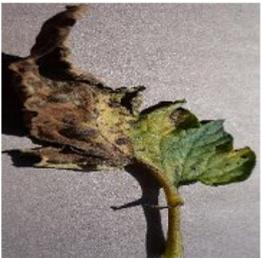 | 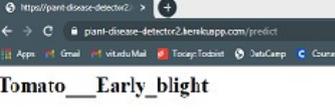 Tomato___Early_blight |

**Fig. 5.** Images and outputs generated by system.

## 5 Conclusion

We have successfully developed a computer vision based system for plant disease detection with average 93% accuracy and 0.93 F1 score. Also the proposed system is computationally efficient because of the use of statistical image processing and machine learning model. Table 3 illustrates the overall benefits of our system over the other approaches.

**Table 3.** Comparison of proposed system with other existing systems.

| Author | S. Khirade et Al. (2015) | Shiroop Madiwalar et Al. (2017) | Peyman Moghadam et Al. (2017) | Sharath D. M. et Al. (2019) | Garima Shrestha et Al. (2020) | Proposed Method |
|---|---|---|---|---|---|---|
| Algorithm | Digital image processing and BPNN | Digital Image processing and SVM | Hyperspectral imaging and SVM | Digital image processing | CNN | Digital image processing and random forest classifier |
| Accuracy | - | 83.34% | 93% | - | 88.80% | 93% |
| Computationally efficient | ✗ | ✓ | ✗ | ✓ | ✗ | ✓ |
| Specialized hardware requirement | ✗ | ✗ | ✓ | ✗ | ✗ | ✗ |

We can observe that our technique is accurate and efficient compared with other systems. Also it won't require a specialized hardware, makes it cost effective solution.